\pdfoutput=1

\documentclass[11pt]{article}

\usepackage{acl}

\usepackage{times}
\usepackage{latexsym}
\usepackage{comment}
\usepackage{amsmath}
\usepackage{xcolor}
\usepackage{graphicx} 
\usepackage{url}
\usepackage{booktabs, multirow, makecell}
\usepackage{balance}
\usepackage{fancyhdr}
\usepackage[boxed]{algorithm2e}

\usepackage[T1]{fontenc}

\usepackage[utf8]{inputenc}

\usepackage{microtype}

\title{Improving Reading Comprehension Question Generation with Data Augmentation and Overgenerate-and-rank}

\author{Nischal Ashok Kumar$^1$, Nigel Fernandez$^1$, Zichao Wang$^2$, Andrew Lan$^1$ \\
  University of Massachusetts Amherst$^1$, Adobe Research$^2$ \\
  \texttt{\{nashokkumar, nigel, andrewlan\}@cs.umass.edu, jackwa@adobe.com}
  }

\begin{document}
\maketitle
\begin{abstract}
Reading comprehension is a crucial skill in many aspects of education, including language learning, cognitive development, and fostering early literacy skills in children. Automated answer-aware reading comprehension question generation has significant potential to scale up learner support in educational activities. One key technical challenge in this setting is that there can be multiple questions, sometimes very different from each other, with the same answer; a trained question generation method may not necessarily know which question human educators would prefer. To address this challenge, we propose 1) a data augmentation method that enriches the training dataset with diverse questions given the same context and answer and 2) an overgenerate-and-rank method to select the best question from a pool of candidates. We evaluate our method on the FairytaleQA dataset, showing a 5\% absolute improvement in ROUGE-L over the best existing method. We also demonstrate the effectiveness of our method in generating harder, ``implicit'' questions, where the answers are not contained in the context as text spans. 

\end{abstract}


\section{Introduction}

Reading comprehension is crucial in assessing students' language learning ability and complex reasoning skills. Comprehending and interpreting stories such as fairy tales, with specific emphasis on narratives, foster early intellectual and literacy development in children~\cite{sim2014shared, lynch2008development}. 
Asking suitable educational-focused questions can help students understand the context of the fairy tales better and inspire their interests~\cite{ganotice2017enhancing, zevenbergen2003dialogic, xu2021same}. However, constructing suitable questions at scale is hard since it is both time intensive and cognitively challenging~\cite{golinkoff2019language}. Researchers have developed models that can automatically generate questions or question-answer pairs to meet the demand for a large pool of relevant questions~\cite{kurdi2020systematic, yao-etal-2022-ais}. These advances can potentially facilitate the development of artificial intelligence (AI)-supported learning platforms to help students develop reading comprehension skills~\cite{zhang2022storybuddy}. 

Prior work on question generation in educational applications can be broadly classified into two categories: \emph{answer-aware}, which is the focus of our current work, and \emph{answer-unaware} (see~\citet{dugan2022feasibility} for a feasibility study), depending on whether the desired answer is given or not. For answer-aware question generation, the goal is to build an AI-based system to generate a question given both the context and the answer~\cite{wang2018qg}. The context can be any text segment, from a few sentences to a possibly long document, that provides background information on which the question is grounded in. The answer is a short span of text that is either part of the context (explicit) or not part of the context but can be inferred from the context (implicit). More specifically, in answer-aware question generation, the question generation system is trained using the context-answer pairs as input and the question as the output~\cite{yao-etal-2022-ais}. 
See Section~\ref{sec:rw} for a detailed discussion on related work. 

A key challenge in answer-aware question generation is that there are often multiple relevant questions for a given context-answer pair. Existing question generation systems are limited in identifying which questions human educators would prefer from multiple relevant ones. Table \ref{tab:multiple_questions} shows an example context-answer pair from the FairytaleQA dataset~\cite{xu-etal-2022-fantastic} with four relevant questions that can be answered by ``a lovely dinner'', the given answer. The first and second questions focus on describing the setting of the context framed using the object (table) and the subject (Tom and Hunca), respectively. The third question adds a causal element inquiring about the cause of Tom and Hunca's emotion. The fourth question is predictive in nature, asking about an event which can be inferred from the context.

Selecting the top question from multiple relevant and diverse question candidates is challenging. For a question generation system to perform this challenging task, it needs to be able to both generate diverse and valid question candidates and also accurately rank and select the top question. To generate diverse question candidates, a question generation system needs to be trained on multiple different relevant questions for a given context-answer pair. To accurately select the top question, a question generation system needs to learn to rank the question candidates by matching the preferences of human educators. We incorporate both of these ideas in our proposed methods in this work.



\subsection{Contributions}
In this paper, we detail two novel methods to improve the robustness of automated answer-aware reading comprehension question generation. We validate their effectiveness through both quantitative and qualitative experiments on the FairytaleQA dataset~\cite{xu-etal-2022-fantastic}; we make our implementation publicly available.\footnote{The code for the paper can be found at: \url{https://github.com/umass-ml4ed/question-gen-aug-ranking}}
Built on top of a Flan-T5 \cite{chung2022scaling} fine-tuning backbone, our contributions are summarized as follows: 
\begin{itemize}
    \item We propose a \textbf{data augmentation} method to augment the training set with synthetically generated diverse and relevant questions. Specifically, we prompt a larger language model, OpenAI Codex~\cite{chen2021evaluating}, to first generate a diverse question pool and then filter out questions that are inconsistent with the given context-answer pair using a question-answering model. 
    \item We propose an \textbf{overgenerate-and-rank} method to rank multiple generated question candidates for the given context-answer pair. Specifically, we fine-tune a separate BERT-based model by optimizing a distribution matching objective to learn which questions are more preferable to human educators and use the model to rank them. 
    \item We conduct extensive experiments to validate the effectiveness of our methods. 
    Our best method achieves a 5\% absolute increase in the ROUGE-L score over the best existing baseline~\cite{xu-etal-2022-fantastic}. 
    We also observe that 1) the data augmentation method can be used to balance questions of different types in the training data and 2) the overgenerate-and-rank method is particularly effective at generating harder questions, i.e., those with answers not explicitly present in the context as text spans.
\end{itemize}

\begin{table}
\small
\centering
\begin{tabular}{p{0.15\linewidth} p{0.73\linewidth}}

\toprule
Context & Tom Thumb and Hunca Munca went up-stairs and peeped into the dining room. Then they squeaked with joy. Such a lovely dinner was laid out upon the table \ldots \\
\midrule

Answer & a lovely dinner\\
\midrule

\multirow{6}{*}{Questions}
& 1. What was laid upon the table? \\
& 2. What did Tom and Hunca see in the dining room? \\
& 3. What made Tom and Hunca squeak with joy? \\
& 4. What will Tom and Hunca enjoy eating in the dining room? \\ 
\bottomrule

\end{tabular}
\caption{Example context-answer pair from the FairytaleQA dataset with multiple valid questions.}
\label{tab:multiple_questions}
\end{table}

\section{Related Work}
\label{sec:rw}

\subsection{QA Datasets on Narratives}
There have been several works proposing QA and QG datasets of educational importance. NarrativeQA~\cite{kovcisky2018narrativeqa} requires students to answer questions written by crowd workers based on books or movie scripts. 
TellMeWhy~\cite{lal-etal-2021-tellmewhy} is another dataset that contains only ``why'' based questions that need additional information not directly present in the text to be answered. A recent and popular dataset to facilitate assessment and training of students' narrative comprehension skills is the FairtytaleQA~\cite{xu-etal-2022-fantastic} dataset. FairtytaleQA contains question-answer pairs written by education experts on fairy tale stories obtained from Project Gutenberg\footnote{https://www.gutenberg.org}. FairtytaleQA is composed of questions focusing on several narrative elements.
We validate the effectiveness of our question generation methods with extensive experiments on FairtytaleQA.

\subsection{Question Generation}
There are several works on question generation for reading comprehension.
%
%
~\citet{stasaski2021automatically} and~\citet{zou2022automatic} propose question generation methods based on causal relations and unsupervised learning, respectively. However, their methods are focused on very specific questions and are thus not generalizable. In contrast, our work focuses on a broad variety of questions covering different narrative elements in reading comprehension.
~\citet{rathod2022educational} proposes to generate multiple semantically similar but lexically diverse questions for a given answer. However, their work is limited to generating only two questions per answer. In contrast, our approach is capable of generating multiple diverse and relevant questions, along with a ranking method to select the best question aligned with human educator preferences.
%
%
Recent work on the FairytaleQA dataset develops event-based question generation methods~\cite{zhao-etal-2022-educational,xu2022nece}. However, their results are reported on only a small subset of attributes: action, causal relationship, and outcome resolution.
In contrast, we report our results over all attributes on the complete FairytaleQA dataset and compare with the current state-of-the-art baseline.
%
%
~\citet{yuan2022selecting} propose a prompt-based question generation method that leverages large language models (LM) like GPT-3. However, these black-box LMs have limited API only access. 
In contrast, our method uses open-source language models to achieve competitive results.
%
%
The FairtytaleQA dataset paper~\cite{xu-etal-2022-fantastic} proposes the current state-of-the-art question generation method by fine-tuning the BART~\cite{lewis-etal-2020-bart} LM to generate the ground truth question given the input context-answer pair. Improving upon LM fine-tuning, we propose two question generation methods for increased robustness, data augmentation and overgenerate-and-rank, which are able to both generate diverse and valid question candidates and also accurately rank and select the top question aligned with human educator preference.


\section{Methodology} \label{sec:method}

In this section, we first introduce the problem setup for question generation on FairytaleQA~\cite{xu-etal-2022-fantastic}. We then detail our question generation approach, building upon the baseline of fine-tuning a language model, by adding our data augmentation method to augment the training set with diverse questions, followed by our over-generate-and-rank method to select the top question from the diverse question candidates generated.

\subsection{Problem Formulation and Dataset Details}
\label{subsection:problem_dataset}
FairytaleQA~\cite{xu-etal-2022-fantastic} is a popular dataset for both question answering and question generation in the education community supporting narrative comprehension, targeting students from kindergarten to eighth grade. Written by education experts, FairytaleQA contains $10,580$ question-answer pairs $(q_i, a_i)$, indexed by $i$, from $278$ classical fairytale stories. Each question-answer pair is sourced from a section of a story referred to as the context $c_i$. The goal for a trained question generation model is to generate the ground truth question $q_i$ conditioned on the input context-answer pair $(c_i, a_i)$. 

Question-answer pairs in FairytaleQA can be categorized in two major ways: 1) by attributes and 2) by the source of answers. In attribute categorization, question-answer pairs capture seven different narrative elements or relations, referred to as attributes, which are character, setting, action, feeling, causal relationship, outcome resolution, and prediction. Orthogonal to the previous categories, questions can also be categorized by whether the answer span is explicitly contained within the context or is implicit and need to be inferred from the context. Explicit questions capture specific story facts while implicit questions require summarization and inference skills. FairytaleQA is imbalanced with respect to question attributes, with action and causal relationship questions accounting for 60\% of the dataset. Our data augmentation method helps balance questions of different attributes.

\subsection{Language Model Fine-tuning} \label{sec:fine-tune_lm}
We first describe our LM fine-tuning approach for question generation. We use a pre-trained Flan-T5~\cite{chung2022scaling} model as our base LM for question generation. We also tried using vanilla T5~\cite{raffel2020exploring} and GPT-2~\cite{radford2019language} as our base LM which gave a comparable but lower performance, possibly because Flan-T5 is instruction fine-tuned 
on a large number of tasks relevant to both QA and QG. Therefore, for simplicity of exposition, we detail our question generation methods using Flan-T5 as the base LM. We construct the input using a combination of the context $c_i$ and answer $a_i$ with the following template: \texttt{Generate question given context and answer: Context:} $c_i$ \texttt{Answer:} $a_i$.

Let $\theta$ represent the LM parameters to be learned. We fine-tune our LM over all context-answer pairs $(c_i, a_i)$ to generate the corresponding ground truth question $q_i$ using a language modeling objective. The language modeling objective is the negative log-likelihood of generating the ground truth question calculated at the token level. The objective $\mathcal{L}_i(\theta)$ for the $i$\textsuperscript{th} training sample is given by:
\begin{align}
\mathcal{L}_i(\theta) = - \sum_{t} \log{P(q_{i,t} | c_i, a_i, q_{i,<t})} \label{eq:loss_objective}
\end{align}
where $q_{i,t}$ is the $t$\textsuperscript{th} token of question $q_i$ and $q_{i,<t}$ refers to all tokens preceding the $t$\textsuperscript{th} token. Our finetuning objective is the sum of this loss across all training questions. 
\subsection{Data Augmentation} 
\label{sec:data_aug}

\begin{figure*}
\includegraphics[width=\linewidth]{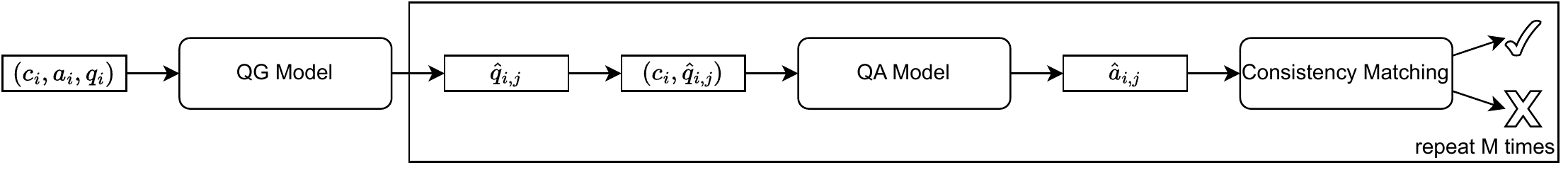}
\caption{Our automated data augmentation method to enrich the training set with diverse and relevant questions for each context-answer-question triplet.} 
\label{fig:data_augmentation}
\end{figure*}

For a question generation system to be robust in selecting the best question for context-answer pairs with multiple relevant questions, it must first be able to generate diverse and suitable question candidates for a given context-answer pair. Moreover, education experts who created the FairytaleQA dataset followed the pattern of first reading the context, then writing a question, and finally writing the answer. This process implies that there could often be multiple valid questions associated with the same context-answer pair in addition to the ground-truth question, which can be used to augment the dataset (as seen in Table~\ref{tab:multiple_questions}).
Therefore, we propose an automated data augmentation method to enrich the training set with diverse and relevant questions for each context-answer-question triplet.
We prompt a larger LM, OpenAI Codex~\cite{chen2021evaluating}, in an in-context prompting fashion~\cite{brown2020language} to first generate diverse questions for each context-answer pair and then filter out unsuitable questions with consistency matching; we detail both steps below.  

\paragraph{Synthetic Data Generation.} 
We first generate synthetic data, i.e., $M=4$ diverse question candidates $\{\hat{q}_{i,1}, \ldots, \hat{q}_{i,M}\}$ for each context-answer-question triplet $(c_i, a_i, q_i)$ using the OpenAI Codex LM~\cite{chen2021evaluating} in an in-context prompting fashion.
%
We construct the in-context prompt by randomly selecting five context-answer-question triplets from the training set with the same attribute as the target context-answer-question triplet to augment. We then append the target triplet followed by the prompt: ``Another question with the same answer is''. These examples help Codex to adapt to the style of questions written by education experts. We use nucleus sampling~\cite{Holtzman2020The} to generate synthetic questions with a p value of $0.9$ and temperature of $0.8$ to ensure diversity. 





\paragraph{Consistency Matching.}
Since there is no guarantee that the generated questions are faithful and match the context-answer pair, we filter out \emph{inconsistent} questions using a consistency matching criterion inspired from~\citet{wang-etal-2021-math}.
A generated question is consistent with respect to its input context-answer-question triplet if the answer of the generated question is the same as (or similar to) the input ground-truth answer. This consistency criterion enables us to include diverse yet consistent synthetic questions to augment the ground-truth questions during training. 

To obtain the answer of a generated question, we again use Codex in an in-context prompting fashion with a subtle change in the prompt. We use the same five in-context examples of context-answer-question triplets taken from the same attribute as the target context-answer-question being augmented. However, we change the earlier context-answer-question pattern suitable for question generation and reformulate in the order of context-question-answer appropriate for question answering.
We denote the answer to the generated question $\hat{q}_{i,j}$ as $\hat{a}_{i,j}$. We use greedy decoding since we need the single best answer. We observe that comparing the similarity of this obtained answer generated by Codex to the ground truth answer $a_i$ written by human education experts can sometimes exclude consistent synthetic questions incorrectly. We alleviate this issue by obtaining another reference answer to compare with; we prompt Codex in an in-context fashion to obtain the answer to the ground truth question $q_i$, which we denote as $\bar{a}_i$. Note that $\bar{a}_i$ could be different from the ground truth answer $a_i$ as shown in an example in Table \ref{tab:gen_question} in the Supplementary Material.

To check consistency, we measure the similarity between $\hat{a}_{i,j}$ and both $a_i$ and $\bar{a}_i$ using the ROUGE-1 F1 score~\cite{lin2004rouge}. If either similarity is greater than a threshold of $0.5$, we include the context-answer-synthetic question triplet $(c_i, a_i, \hat{q}_{i,j})$ in our augmented training set. We outline our method in Figure~\ref{fig:data_augmentation} and also in Algorithm~\ref{algo:data_augmentation} in the Supplementary Material.

\subsection{Overgenerate-and-Rank} 

\begin{figure}
\includegraphics[width=\linewidth]{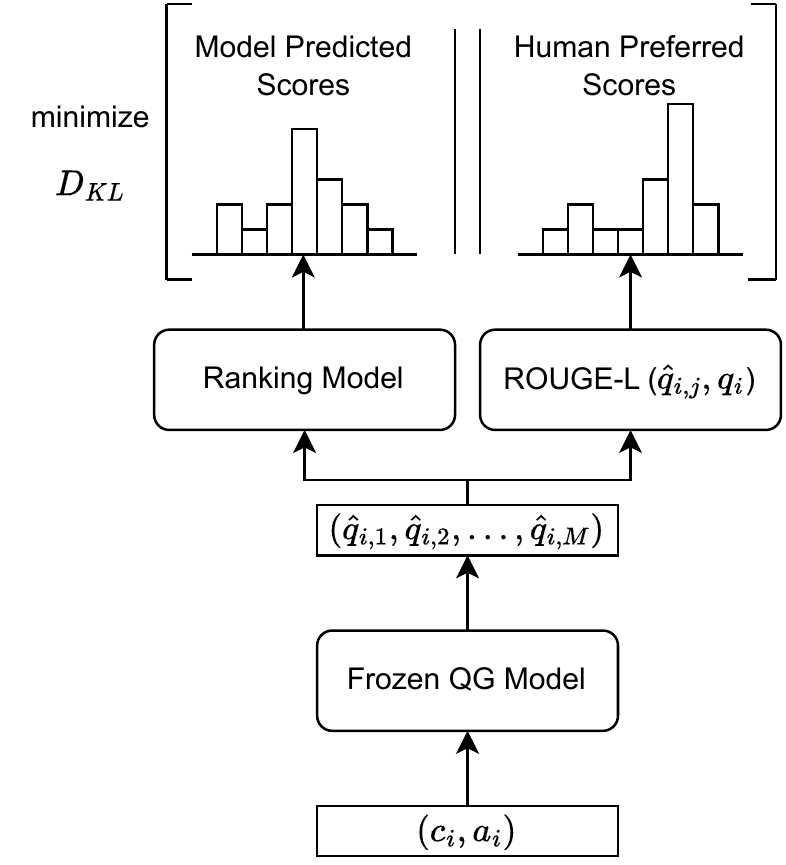}
\caption{The training process of the ranking model used in our overgenerate-and-rank method with distribution matching-based ranking.} 
\label{fig:overgenerate_and_rank}
\end{figure}

Selecting the top question preferable to human educators from multiple relevant and diverse question candidates for the given context-answer pair is hard. We propose an overgenerate-and-rank method which first overgenerates several question candidates for each context-answer pair using the fine-tuned model (as described in Section \ref{sec:fine-tune_lm}). We use various decoding strategies, including nucleus sampling~\cite{Holtzman2020The} and contrastive search~\cite{su2022a} to ensure diversity. We then rank these generated questions based on a criterion. We use two kinds of ranking methods, perplexity-based ranking and distribution matching-based ranking, which we detail below.

\paragraph{Perplexity-based Ranking.} 
In this ranking method, we use perplexity as a metric to rank the generated questions. The perplexity of a language model given a question measures the uncertainty of generating the question under that language model. 
The lower the perplexity of a question, the more probable is the question according to the language model. 
We first overgenerate $K=10$ questions for the given context-answer pair using nucleus sampling or contrastive search. We then compute the perplexity of these questions given the fine-tuned language model. We then select the question with the lowest perplexity as the best question for the given context-answer pair. 

\paragraph{Distribution Matching-based Ranking.} 
In this ranking method, we fine-tune a separate language model to rank the overgenerated question candidates by  predicting scores over these generated questions with a similar distribution to the ROUGE-L scores between the generated questions and the ground truth question. This distribution matching objective encourages the ranking language model to associate higher scores with questions similar to the ground truth question written by human education experts. We select the question with the highest score predicted by the ranking model as the best question for the given context-answer pair.
Our method inspired from~\cite{shi2023replug} trains a ranking language model to minimize the KL divergence~\cite{joyce2011kullback} between the distribution of the model-predicted scores over the generated questions and the distribution of ROUGE-L scores computing similarity of the generated questions to the human educator-written ground truth question. 
We outline the training process of the ranking model in Figure~\ref{fig:overgenerate_and_rank}.

More specifically, we use a pre-trained ConvBERT~\cite{jiang2020convbert} model as our ranking language model. We use a combination of the given context-answer pair and the generated question to rank as input to the  model. We feed the \texttt{[CLS]} embedding vector to a learnable linear layer during fine-tuning.
For the $i\textsuperscript{th}$ training question, $P_{\phi}(\hat{q}_i) \in [0,1]^K$ denotes the probability distribution of the model-predicted scores for generated questions and $R(\hat{q}_i, q_i) \in [0,1]^K$ denotes the probability distribution of the ROUGE-L scores between the generated questions and the ground-truth question. Equation \ref{eq:kl_loss} shows the fine-tuning objective of the ranking language model to mimize the KL divergence between the model-predicted score distribution and the ROUGE-L score distribution. The softmax in equation \ref{eq:lm_softmax} computes the distribution of the model-predicted scores where $\phi(\hat{q}_{i, j}, c_i, a_i)$ denotes the score predicted by the ranking language model for the $j\textsuperscript{th}$ generated question $\hat{q}_{i, j}$ corresponding to the $i\textsuperscript{th}$ context-answer pair $(c_i, a_i)$. The softmax in equation \ref{eq:rouge_softmax} computes the probability distribution of the ROUGE-L scores where $r(\hat{q}_{i, j}, q_i)$ denotes the ROUGE-L score between the $j^{\text{th}}$ generated question $\hat{q}_{i, j}$ and the ground-truth question $q_i$. The hyperparameters $\alpha_P$ and $\alpha_R$ control the temperature of the softmax over the model-predicted scores and the ROUGE-L scores, respectively. The optimization problem is formally written as: 
\begin{align}
\text{minimize}_\phi \quad \frac{1}{N} & \sum_{i}^N KL (P_{\phi}(\hat{q_i}) || R(\hat{q_i}, q_i)), \label{eq:kl_loss}\\
\text{where} \; [P_{\phi}(\hat{q_i})]_j & = \frac{\exp{\alpha_P.\phi(\hat{q}_{i, j}, c_i, a_i)}}{\sum_j \exp{\alpha_P.\phi(\hat{q}_{i, j}, c_i, a_i)}}, \label{eq:lm_softmax} \\
[R(\hat{q_i}, q_i)]_j & = \frac{\exp{\alpha_R.r(\hat{q}_{i, j}, q_i)}}{\sum_j \exp{\alpha_R.r(\hat{q}_{i, j}, q_i)}}. \label{eq:rouge_softmax}
\end{align}


\section{Experimental Evaluation} 

In this section, we 
describe the experimental setup to validate the effectiveness of our question generation methods.

\subsection{Metrics and Baselines}

To compare with prior work~\cite{xu-etal-2022-fantastic}, we use the ROUGE-L F1 score~\cite{lin2004rouge} (referred to as ROUGE-L) to evaluate the quality of generated questions. 
We compare our question generation methods to the existing state-of-the-art baseline~\cite{xu-etal-2022-fantastic} which fine-tunes a BART LM~\cite{lewis-etal-2020-bart} to generate the ground truth question conditioned on the given context-answer pair.

\subsection{Implementation Details}
We use a pre-trained Flan-T5-Large model~\cite{chung2022scaling} with $770$M parameters as our base LM for question generation; all implementation was done using the HuggingFace~\cite{wolf-etal-2020-transformers} transformers library. We fine-tune the base LM for 10 epochs with early stopping on the validation loss using the AdamW~\cite{Loshchilov2017DecoupledWD} optimizer with a learning rate of 3e-4 and a batch size of 8.
Each epoch takes $20$ minutes on a single NVIDIA A100 GPU.

FairytaleQA is imbalanced with respect to question attributes, with action and causal relationship accounting for 60\% of the dataset. Our data augmentation method generates around $2500$ synthetic questions over only the minority attributes: character, setting, feeling, outcome resolution, and prediction, to balance the training set. We fine-tune our base LM with the same setup described before on the augmented training set using a weight $\lambda$ for the loss objective (see Equation~\ref{eq:loss_objective}) with original human educator-written questions and a different weight $1-\lambda$ for synthetic questions. Through a grid search, we find that setting $\lambda=0.8$ results in the best performance.

Our overgenerate-and-rank method generates question candidates using contrastive search~\cite{su2022a} (top-k of 4, $\alpha$ penalty of 0.6) and nucleus sampling~\cite{Holtzman2020The} (top-p of 0.9, temperature of 1) for perplexity-based ranking and distribution matching-based ranking, respectively. Through a grid search, we find that setting the softmax temperature hyperparameters as $\alpha_P=1e-3$ and $\alpha_R=1e-2$ results in the best performance.


\section{Results and Discussion} \label{sec:results}
\paragraph{Overall Performance.}
We report the average ROUGE-L across all test questions in the FairytaleQA dataset for all question generation methods in Table~\ref{tab:results_main}. The choice of the base language model is key when fine-tuning language models for question generation; fine-tuning Flan-T5 provides a significant improvement of $3.7\%$ over the current state-of-the-art baseline of fine-tuning BART~\cite{xu-etal-2022-fantastic}, possibly because Flan-T5 is instruction fine-tuned on a large number of tasks relevant to both question answering and question generation. Our data augmentation method, which enriches the training set with diverse questions, further improves performance by $0.25\%$ over fine-tuning Flan-T5 on the original training set. Among our overgenerate-and-rank methods, perplexity-based ranking and distribution matching-based ranking provide a $0.5\%$ and $1.4\%$ improvement over fine-tuning Flan-T5, respectively. Overall, our best method, distribution matching-based ranking method, provides a $5\%$ absolute improvement over the current state-of-the-art BART baseline. 
This significant improvement shows that our data augmentation and overgenerate-and-rank methods are effective at making question-generation systems more robust, which results in better questions being generated. 
We also experiment with combining our data augmentation and overgenerate-and-rank methods. However, perhaps surprisingly, this combination does not lead to significant improvement in performance. We think that this result is possibly due to synthetic questions being too diverse in many cases with respect to the ground truth question. Therefore, controlling the diversity of synthetic questions for better alignment with those written by human educators is an important direction for future work.



\paragraph{Performance Stratified by Question Category.} To gain more insight into the performance of our question generation methods, we also report the average ROUGE-L over test questions in the explicit and implicit categories. 
For the harder implicit questions with answers not explicitly included in the context as text spans, our data augmentation and distribution matching-based ranking methods improve performance by $1.2\%$ and $2.3\%$ over fine-tuning Flan-T5, respectively. 
This significant performance improvement shows that our data augmentation and overgenerate-and-rank methods are well-suited for harder question generation tasks, especially when given an answer that needs to be inferred from the context, for which the ground-truth questions are already highly diverse. 

\begin{table}[tp]
\centering
\scalebox{0.95}{
\begin{tabular}{l c c c}
\toprule
\multirow{2}{*}{Method} & \multicolumn{3}{c}{Questions}\\
\cmidrule{2-4}
& All & Explicit & Implicit\\
\midrule
\begin{tabular}[c]{@{}l@{}}BART  \\~\cite{xu-etal-2022-fantastic} \end{tabular}                   & 0.5270          & -                                                              & -                                                              \\ \midrule
\begin{tabular}[c]{@{}l@{}}Flan-T5\end{tabular}           & 0.5639                                                             & 0.5998                                                              & 0.4571                                                              \\ \midrule
\begin{tabular}[c]{@{}l@{}}Data \\ Augmentation\end{tabular}                & 0.5664                                                             & 0.5994                                                              & 0.4682                                                              \\ \midrule
\begin{tabular}[c]{@{}l@{}}Perplexity-based \\ Ranking\end{tabular}               & 0.5689                                                             & 0.6057                                                              & 0.4591                                                              \\ \midrule
\begin{tabular}[c]{@{}l@{}}Distribution \\ Matching-based \\ Ranking\end{tabular} & \textbf{0.5778}                                                    & \textbf{0.6107}                                                     & \textbf{0.4798}                                                     \\ 
\bottomrule
\end{tabular}}
\caption{Experimental results on the FairytaleQA dataset in ROUGE-L (higher is better). Our methods significantly outperform existing baselines.}
\label{tab:results_main}
\end{table}

\paragraph{Data Augmentation Variants.}
We report ROUGE-L scores for several variants of our data augmentation method in Table \ref{tab:aug_results} in the Supplementary Material. FairytaleQA is imbalanced with respect to question attributes, with action and causal relationship accounting for $60\%$ of the dataset. Augmenting all questions across all attributes results in a drop in performance. This observation validates our best data augmentation method, which is to generate synthetic questions for only the minority attributes: character, setting, feeling, outcome resolution, and prediction, to balance the training set. Moreover, fine-tuning Flan-T5 by weighting the human educator-written questions and synthetically-generated questions differently further improves performance.

\paragraph{Different Decoding Strategies.}
We report ROUGE-L scores for our overgenerate-and-rank methods combined with different choices of decoding strategy for overgeneration: greedy, nucleus sampling, and contrastive search, in Table~\ref{tab:results_rank} in the Supplementary Material. We compare perplexity-based ranking and two variants of distribution matching-based ranking trained on questions generated by nucleus sampling and contrastive search, respectively. We see that there is no single best decoding strategy that works across all ranking methods. We also observe that using the same decoding strategy for overgenerating candidate questions for both training and testing of the ranking method might not provide the best performance. For example, the distribution matching-based ranking method trained on questions generated by contrastive search works best at test time by ranking questions generated by nucleus sampling.



\begin{table*}[tp]
\small
\centering
\begin{tabular}{{p{0.22\linewidth} p{0.73\linewidth}}}
\toprule

Context & \ldots and when they had finished the little grey old man said to the dullhead: ``Now I will bring you luck, because you have a kind heart and are willing to share what you have with others. there stands an old tree; cut it down, and amongst its roots, you'll find something.'' \ldots \\ 
\midrule

Answer & cut down an old tree.\\ 
\midrule

Ground truth question & What did the man tell dullhead to do? \\

\midrule
Flan-T5 & What did dullhead need to do to find something?\\ \midrule

Data Augmentation & What did the little grey old man tell dullhead to do?\\ \midrule

\multirow{5}{*} {Perplexity-based Ranking} & 1. What did the little man tell dullhead to do because he was willing to share what he had?\\
& 2. What did the little man tell dullhead to do because he wanted to find something?\\
& 3. What will dullhead do after he has eaten and drank the cake and beer?\\
& 4. What will dullhead do to find something?\\ 
& 5. What will dullhead do when he meets the grey old man?\\ 
\midrule

\multirow{5}{*}{\shortstack[l]{Distribution\\ Matching-based Ranking}} 
& 1. What did the grey old man ask dullhead to do?\\ 
& 2. What did the little grey old man say he wanted dullhead to do?\\ 
& 3. What did the little man tell dullhead to do because he was willing to share what he had?\\ 
& 4. What did the little man tell dullhead to do because he wanted to find something?\\ 
& 5. What will dullhead need to do?\\ 
\bottomrule

\end{tabular}
\caption{Qualitative analysis with an example input context-answer-question from the FairytaleQA dataset and question generated by our methods. Both data augmentation and overgenerate-and-rank improve diversity among the generated questions, which makes question generation more robust.}
\label{tab:qualitative_analysis}
\end{table*}

\subsection{Qualitative Analysis}
\paragraph{Analysis of Questions Generated.}
We provide a qualitative analysis of our question generation methods on an example context-answer-question triplet from the test set of FairytaleQA in Table \ref{tab:qualitative_analysis}. We observe that there are multiple relevant questions with different linguistic style and structure for the example context-answer pair; among them, our question generation methods need to generate the human educator-written ground truth question, ``What did the man tell dullhead to do?''. Our fine-tuned Flan-T5 model generates a plausible but vague question excluding the subject of the context, ``the old man'', that is not very similar to the ground truth question, possibly due to limitations of greedy decoding. Our data augmentation method generates a much better question that is similar in structure and style to the ground truth question, which suggests that training on diverse questions is effective.

We also show the top five questions among the candidates, ranked by our overgenerate-and-rank methods. Our perplexity-based ranking method improves upon the fine-tuned Flan-T5 model by matching the structure of the ground truth question, ``What did the man tell dullhead \ldots'', but favors longer questions with more context information than the human educator-written question. Our distribution matching-based ranking method performs best by matching both the structure and style of the ground truth question. This example demonstrates that ranking methods trained on actual human preference information can be effective at identifying human-like questions among diverse candidates.
\paragraph{Error Analysis.}
We randomly select 30 context-answer pairs from the FairytaleQA test set with low ROGUE-L scores (less than 0.2) and investigate the questions generated by our best method, distribution matching-based ranking, and analyze why it does not perform well in these cases. We identify three main error types and list them in Table~\ref{tab:qualitative_error_types} in the Supplementary Material, with corresponding examples containing the input context-answer pair, the ground truth question, and the best generated question. The three main error types are: 1) character coreference resolution, 2) out-of-context ground-truth questions, and 3) multiple evidence angles in the context.

The first two error types are beyond our control but the third type suggests that our methods have plenty of room for improvement. 
Errors of type character coreference resolution can occur when an input context has multiple characters and coreferences. In the first example, ``self'' is used as a complex coreference and confuses the question generation method.
Errors of type out-of-context ground-truth questions can occur for ground-truth questions using information present outside the context the model sees as input. These ground-truth questions are human errors often referring to named entities present in other sections of the same story but not included in the input context. In the second example, the ground truth question refers to the character ``Ian'' who is not present in the context; the generated question uses the reference of ``fisher's son'' that is has access to in the given context. 
Errors of type multiple evidence angles can occur when the input context discusses different aspects of an answer. In the third example, the event of ``Norseman invasion'' in the answer could have questions related to either its cause, ``people being wicked'', or its timeline, ``happening after the two Countesses fled to Scotland''. As a result, among the top decoder output questions, there are none that discusses the latter, which is contained in the ground-truth question. Therefore, it is important to develop methods that can take all possible question angles into account during decoding. 


\section{Conclusions and Future Work}

In this paper, we proposed methods for improving automated answer-aware reading comprehension question generation by generating diverse question candidates and ranking them to align with human educator preferences.
First, we proposed a data augmentation method that augments the training dataset with diverse questions obtained from a larger language model. Second, we proposed an overgenerate-and-rank method with two choices of ranking criterion, perplexity-based ranking and distribution matching-based ranking. The latter learns to rank the generated candidate questions to select ones that are closer to human-written questions.
We conducted extensive experiments on the FairytaleQA dataset to validate the effectiveness of our methods showing that our best method provides an absolute improvement of 5\% in ROUGE-L over the current state-of-the-art on this dataset. We also showed that our methods are significantly better than baselines in generating harder questions whose answers are not directly present in the context as text spans and have to be inferred. 

There are several directions for future work. First, we can experiment with other data augmentation methods, e.g., by fine-tuning the base language model by weighting synthetically-generated questions according to their ROUGE-L scores with respect to the ground truth question. Second, we can explore the use of chain-of-thought~\cite{wei2022chain} or self-ask~\cite{press2022measuring} to prompt the large language model in our data augmentation method. Third, we can experiment with other ranking objectives, such as ones using the Bradley-Terry model~\cite{bradley1952rank} or ones using reinforcement learning with human feedback framework~\cite{ziegler2019fine}, to select the best questions that are aligned with human preference. Fourth, we can apply our methods to other question generation scenarios that require reasoning, such as logistical questions in online course discussion forums \cite{zylich2020exploring}, to help instructors anticipate common student questions.

\section*{Acknowledgements}
We thank the anonymous reviewers for their helpful comments. We thank the Learning Agency Lab for organizing the Quest for Quality Questions challenge\footnote{https://www.thequestchallenge.org} which inspired our initial work. We also thank Alexander Scarlatos and Naiming Liu for helpful discussions around this work. The authors also thank the NSF (under grants 1917713, 2118706, 2202506, 2215193, 2237676) for partially supporting this work.

\bibliographystyle{acl_natbib}
\bibliography{custom}

\begin{thebibliography}{40}
\expandafter\ifx\csname natexlab\endcsname\relax\def\natexlab#1{#1}\fi

\bibitem[{Bradley and Terry(1952)}]{bradley1952rank}
Ralph~Allan Bradley and Milton~E Terry. 1952.
\newblock Rank analysis of incomplete block designs: I. the method of paired
  comparisons.
\newblock \emph{Biometrika}, 39(3/4):324--345.

\bibitem[{Brown et~al.(2020)Brown, Mann, Ryder, Subbiah, Kaplan, Dhariwal,
  Neelakantan, Shyam, Sastry, Askell et~al.}]{brown2020language}
Tom Brown, Benjamin Mann, Nick Ryder, Melanie Subbiah, Jared~D Kaplan, Prafulla
  Dhariwal, Arvind Neelakantan, Pranav Shyam, Girish Sastry, Amanda Askell,
  et~al. 2020.
\newblock Language models are few-shot learners.
\newblock \emph{Advances in neural information processing systems},
  33:1877--1901.

\bibitem[{Chen et~al.(2021)Chen, Tworek, Jun, Yuan, Pinto, Kaplan, Edwards,
  Burda, Joseph, Brockman et~al.}]{chen2021evaluating}
Mark Chen, Jerry Tworek, Heewoo Jun, Qiming Yuan, Henrique Ponde de~Oliveira
  Pinto, Jared Kaplan, Harri Edwards, Yuri Burda, Nicholas Joseph, Greg
  Brockman, et~al. 2021.
\newblock Evaluating large language models trained on code.
\newblock \emph{arXiv preprint arXiv:2107.03374}.

\bibitem[{Chung et~al.(2022)Chung, Hou, Longpre, Zoph, Tay, Fedus, Li, Wang,
  Dehghani, Brahma et~al.}]{chung2022scaling}
Hyung~Won Chung, Le~Hou, Shayne Longpre, Barret Zoph, Yi~Tay, William Fedus,
  Eric Li, Xuezhi Wang, Mostafa Dehghani, Siddhartha Brahma, et~al. 2022.
\newblock Scaling instruction-finetuned language models.
\newblock \emph{arXiv preprint arXiv:2210.11416}.

\bibitem[{Dugan et~al.(2022)Dugan, Miltsakaki, Upadhyay, Ginsberg, Gonzalez,
  Choi, Yuan, and Callison-Burch}]{dugan2022feasibility}
Liam Dugan, Eleni Miltsakaki, Shriyash Upadhyay, Etan Ginsberg, Hannah
  Gonzalez, DaHyeon Choi, Chuning Yuan, and Chris Callison-Burch. 2022.
\newblock A feasibility study of answer-unaware question generation for
  education.
\newblock In \emph{Findings of the Association for Computational Linguistics:
  ACL 2022}, pages 1919--1926.

\bibitem[{Ganotice~Jr et~al.(2017)Ganotice~Jr, Downing, Mak, Chan, and
  Lee}]{ganotice2017enhancing}
Fraide~A Ganotice~Jr, Kevin Downing, Teresa Mak, Barbara Chan, and Wai~Yip Lee.
  2017.
\newblock Enhancing parent-child relationship through dialogic reading.
\newblock \emph{Educational Studies}, 43(1):51--66.

\bibitem[{Golinkoff et~al.(2019)Golinkoff, Hoff, Rowe, Tamis-LeMonda, and
  Hirsh-Pasek}]{golinkoff2019language}
Roberta~Michnick Golinkoff, Erika Hoff, Meredith~L Rowe, Catherine~S
  Tamis-LeMonda, and Kathy Hirsh-Pasek. 2019.
\newblock Language matters: Denying the existence of the 30-million-word gap
  has serious consequences.
\newblock \emph{Child development}, 90(3):985--992.

\bibitem[{Holtzman et~al.(2020)Holtzman, Buys, Du, Forbes, and
  Choi}]{Holtzman2020The}
Ari Holtzman, Jan Buys, Li~Du, Maxwell Forbes, and Yejin Choi. 2020.
\newblock \href {https://openreview.net/forum?id=rygGQyrFvH} {The curious case
  of neural text degeneration}.
\newblock In \emph{International Conference on Learning Representations}.

\bibitem[{Jiang et~al.(2020)Jiang, Yu, Zhou, Chen, Feng, and
  Yan}]{jiang2020convbert}
Zi-Hang Jiang, Weihao Yu, Daquan Zhou, Yunpeng Chen, Jiashi Feng, and Shuicheng
  Yan. 2020.
\newblock Convbert: Improving bert with span-based dynamic convolution.
\newblock \emph{Advances in Neural Information Processing Systems},
  33:12837--12848.

\bibitem[{Joyce(2011)}]{joyce2011kullback}
James~M Joyce. 2011.
\newblock Kullback-leibler divergence.
\newblock In \emph{International encyclopedia of statistical science}, pages
  720--722. Springer.

\bibitem[{Ko{\v{c}}isk{\`y} et~al.(2018)Ko{\v{c}}isk{\`y}, Schwarz, Blunsom,
  Dyer, Hermann, Melis, and Grefenstette}]{kovcisky2018narrativeqa}
Tom{\'a}{\v{s}} Ko{\v{c}}isk{\`y}, Jonathan Schwarz, Phil Blunsom, Chris Dyer,
  Karl~Moritz Hermann, G{\'a}bor Melis, and Edward Grefenstette. 2018.
\newblock The narrativeqa reading comprehension challenge.
\newblock \emph{Transactions of the Association for Computational Linguistics},
  6:317--328.

\bibitem[{Kurdi et~al.(2020)Kurdi, Leo, Parsia, Sattler, and
  Al-Emari}]{kurdi2020systematic}
Ghader Kurdi, Jared Leo, Bijan Parsia, Uli Sattler, and Salam Al-Emari. 2020.
\newblock A systematic review of automatic question generation for educational
  purposes.
\newblock \emph{International Journal of Artificial Intelligence in Education},
  30:121--204.

\bibitem[{Lal et~al.(2021)Lal, Chambers, Mooney, and
  Balasubramanian}]{lal-etal-2021-tellmewhy}
Yash~Kumar Lal, Nathanael Chambers, Raymond Mooney, and Niranjan
  Balasubramanian. 2021.
\newblock \href {https://doi.org/10.18653/v1/2021.findings-acl.53}
  {{T}ell{M}e{W}hy: A dataset for answering why-questions in narratives}.
\newblock In \emph{Findings of the Association for Computational Linguistics:
  ACL-IJCNLP 2021}, pages 596--610, Online. Association for Computational
  Linguistics.

\bibitem[{Lewis et~al.(2020)Lewis, Liu, Goyal, Ghazvininejad, Mohamed, Levy,
  Stoyanov, and Zettlemoyer}]{lewis-etal-2020-bart}
Mike Lewis, Yinhan Liu, Naman Goyal, Marjan Ghazvininejad, Abdelrahman Mohamed,
  Omer Levy, Veselin Stoyanov, and Luke Zettlemoyer. 2020.
\newblock \href {https://doi.org/10.18653/v1/2020.acl-main.703} {{BART}:
  Denoising sequence-to-sequence pre-training for natural language generation,
  translation, and comprehension}.
\newblock In \emph{Proceedings of the 58th Annual Meeting of the Association
  for Computational Linguistics}, pages 7871--7880, Online. Association for
  Computational Linguistics.

\bibitem[{Lin(2004)}]{lin2004rouge}
Chin-Yew Lin. 2004.
\newblock Rouge: A package for automatic evaluation of summaries.
\newblock In \emph{Text summarization branches out}, pages 74--81.

\bibitem[{Loshchilov and Hutter(2017)}]{Loshchilov2017DecoupledWD}
Ilya Loshchilov and Frank Hutter. 2017.
\newblock Decoupled weight decay regularization.
\newblock In \emph{International Conference on Learning Representations}.

\bibitem[{Lynch et~al.(2008)Lynch, Van Den~Broek, Kremer, Kendeou, White, and
  Lorch}]{lynch2008development}
Julie~S Lynch, Paul Van Den~Broek, Kathleen~E Kremer, Panayiota Kendeou,
  Mary~Jane White, and Elizabeth~P Lorch. 2008.
\newblock The development of narrative comprehension and its relation to other
  early reading skills.
\newblock \emph{Reading Psychology}, 29(4):327--365.

\bibitem[{Press et~al.(2022)Press, Zhang, Min, Schmidt, Smith, and
  Lewis}]{press2022measuring}
Ofir Press, Muru Zhang, Sewon Min, Ludwig Schmidt, Noah~A Smith, and Mike
  Lewis. 2022.
\newblock Measuring and narrowing the compositionality gap in language models.
\newblock \emph{arXiv preprint arXiv:2210.03350}.

\bibitem[{Radford et~al.(2019)Radford, Wu, Child, Luan, Amodei, Sutskever
  et~al.}]{radford2019language}
Alec Radford, Jeffrey Wu, Rewon Child, David Luan, Dario Amodei, Ilya
  Sutskever, et~al. 2019.
\newblock Language models are unsupervised multitask learners.
\newblock \emph{OpenAI blog}, 1(8):9.

\bibitem[{Raffel et~al.(2020)Raffel, Shazeer, Roberts, Lee, Narang, Matena,
  Zhou, Li, and Liu}]{raffel2020exploring}
Colin Raffel, Noam Shazeer, Adam Roberts, Katherine Lee, Sharan Narang, Michael
  Matena, Yanqi Zhou, Wei Li, and Peter~J Liu. 2020.
\newblock Exploring the limits of transfer learning with a unified text-to-text
  transformer.
\newblock \emph{The Journal of Machine Learning Research}, 21(1):5485--5551.

\bibitem[{Rathod et~al.(2022)Rathod, Tu, and Stasaski}]{rathod2022educational}
Manav Rathod, Tony Tu, and Katherine Stasaski. 2022.
\newblock Educational multi-question generation for reading comprehension.
\newblock In \emph{Proceedings of the 17th Workshop on Innovative Use of NLP
  for Building Educational Applications (BEA 2022)}, pages 216--223.

\bibitem[{Shi et~al.(2023)Shi, Min, Yasunaga, Seo, James, Lewis, Zettlemoyer,
  and Yih}]{shi2023replug}
Weijia Shi, Sewon Min, Michihiro Yasunaga, Minjoon Seo, Rich James, Mike Lewis,
  Luke Zettlemoyer, and Wen-tau Yih. 2023.
\newblock Replug: Retrieval-augmented black-box language models.
\newblock \emph{arXiv preprint arXiv:2301.12652}.

\bibitem[{Sim and Berthelsen(2014)}]{sim2014shared}
Susan Sim and Donna Berthelsen. 2014.
\newblock Shared book reading by parents with young children: Evidence-based
  practice.
\newblock \emph{Australasian Journal of Early Childhood}, 39(1):50--55.

\bibitem[{Stasaski et~al.(2021)Stasaski, Rathod, Tu, Xiao, and
  Hearst}]{stasaski2021automatically}
Katherine Stasaski, Manav Rathod, Tony Tu, Yunfang Xiao, and Marti~A Hearst.
  2021.
\newblock Automatically generating cause-and-effect questions from passages.
\newblock In \emph{Proceedings of the 16th Workshop on Innovative Use of NLP
  for Building Educational Applications}, pages 158--170.

\bibitem[{Su et~al.(2022)Su, Lan, Wang, Yogatama, Kong, and Collier}]{su2022a}
Yixuan Su, Tian Lan, Yan Wang, Dani Yogatama, Lingpeng Kong, and Nigel Collier.
  2022.
\newblock \href {https://openreview.net/forum?id=V88BafmH9Pj} {A contrastive
  framework for neural text generation}.
\newblock In \emph{Advances in Neural Information Processing Systems}.

\bibitem[{Wang et~al.(2021)Wang, Lan, and Baraniuk}]{wang-etal-2021-math}
Zichao Wang, Andrew Lan, and Richard Baraniuk. 2021.
\newblock \href {https://doi.org/10.18653/v1/2021.emnlp-main.484} {Math word
  problem generation with mathematical consistency and problem context
  constraints}.
\newblock In \emph{Proceedings of the 2021 Conference on Empirical Methods in
  Natural Language Processing}, pages 5986--5999, Online and Punta Cana,
  Dominican Republic. Association for Computational Linguistics.

\bibitem[{Wang et~al.(2018)Wang, Lan, Nie, Waters, Grimaldi, and
  Baraniuk}]{wang2018qg}
Zichao Wang, Andrew~S Lan, Weili Nie, Andrew~E Waters, Phillip~J Grimaldi, and
  Richard~G Baraniuk. 2018.
\newblock Qg-net: a data-driven question generation model for educational
  content.
\newblock In \emph{Proceedings of the fifth annual ACM conference on learning
  at scale}, pages 1--10.

\bibitem[{Wei et~al.(2022)Wei, Wang, Schuurmans, Bosma, Chi, Le, and
  Zhou}]{wei2022chain}
Jason Wei, Xuezhi Wang, Dale Schuurmans, Maarten Bosma, Ed~Chi, Quoc Le, and
  Denny Zhou. 2022.
\newblock Chain of thought prompting elicits reasoning in large language
  models.
\newblock \emph{arXiv preprint arXiv:2201.11903}.

\bibitem[{Wolf et~al.(2020)Wolf, Debut, Sanh, Chaumond, Delangue, Moi, Cistac,
  Rault, Louf, Funtowicz, Davison, Shleifer, von Platen, Ma, Jernite, Plu, Xu,
  Scao, Gugger, Drame, Lhoest, and Rush}]{wolf-etal-2020-transformers}
Thomas Wolf, Lysandre Debut, Victor Sanh, Julien Chaumond, Clement Delangue,
  Anthony Moi, Pierric Cistac, Tim Rault, Rémi Louf, Morgan Funtowicz, Joe
  Davison, Sam Shleifer, Patrick von Platen, Clara Ma, Yacine Jernite, Julien
  Plu, Canwen Xu, Teven~Le Scao, Sylvain Gugger, Mariama Drame, Quentin Lhoest,
  and Alexander~M. Rush. 2020.
\newblock Transformers: State-of-the-art natural language processing.
\newblock In \emph{EMNLP: System Demonstrations}, pages 38--45.

\bibitem[{Xu et~al.(2022{\natexlab{a}})Xu, Isaza, Li, Oloko, Yao, Adebeyi, Hou,
  Peng, and Wang}]{xu2022nece}
Guangxuan Xu, Paulina~Toro Isaza, Moshi Li, Akintoye Oloko, Bingsheng Yao,
  Aminat Adebeyi, Yufang Hou, Nanyun Peng, and Dakuo Wang. 2022{\natexlab{a}}.
\newblock Nece: Narrative event chain extraction toolkit.
\newblock \emph{arXiv preprint arXiv:2208.08063}.

\bibitem[{Xu et~al.(2021)Xu, Wang, Collins, Lee, and Warschauer}]{xu2021same}
Ying Xu, Dakuo Wang, Penelope Collins, Hyelim Lee, and Mark Warschauer. 2021.
\newblock Same benefits, different communication patterns: Comparing children's
  reading with a conversational agent vs. a human partner.
\newblock \emph{Computers \& Education}, 161:104059.

\bibitem[{Xu et~al.(2022{\natexlab{b}})Xu, Wang, Yu, Ritchie, Yao, Wu, Zhang,
  Li, Bradford, Sun, Hoang, Sang, Hou, Ma, Yang, Peng, Yu, and
  Warschauer}]{xu-etal-2022-fantastic}
Ying Xu, Dakuo Wang, Mo~Yu, Daniel Ritchie, Bingsheng Yao, Tongshuang Wu, Zheng
  Zhang, Toby Li, Nora Bradford, Branda Sun, Tran Hoang, Yisi Sang, Yufang Hou,
  Xiaojuan Ma, Diyi Yang, Nanyun Peng, Zhou Yu, and Mark Warschauer.
  2022{\natexlab{b}}.
\newblock \href {https://doi.org/10.18653/v1/2022.acl-long.34} {Fantastic
  questions and where to find them: {F}airytale{QA} {--} an authentic dataset
  for narrative comprehension}.
\newblock In \emph{Proceedings of the 60th Annual Meeting of the Association
  for Computational Linguistics (Volume 1: Long Papers)}, pages 447--460,
  Dublin, Ireland. Association for Computational Linguistics.

\bibitem[{Yao et~al.(2022)Yao, Wang, Wu, Zhang, Li, Yu, and
  Xu}]{yao-etal-2022-ais}
Bingsheng Yao, Dakuo Wang, Tongshuang Wu, Zheng Zhang, Toby Li, Mo~Yu, and Ying
  Xu. 2022.
\newblock \href {https://doi.org/10.18653/v1/2022.acl-long.54} {It is {AI}{'}s
  turn to ask humans a question: Question-answer pair generation for
  children{'}s story books}.
\newblock In \emph{Proceedings of the 60th Annual Meeting of the Association
  for Computational Linguistics (Volume 1: Long Papers)}, pages 731--744,
  Dublin, Ireland. Association for Computational Linguistics.

\bibitem[{Yuan et~al.(2022)Yuan, Wang, Wang, Fine, Abdelghani, Lucas,
  Sauz{\'e}on, and Oudeyer}]{yuan2022selecting}
Xingdi Yuan, Tong Wang, Yen-Hsiang Wang, Emery Fine, Rania Abdelghani, Pauline
  Lucas, H{\'e}l{\`e}ne Sauz{\'e}on, and Pierre-Yves Oudeyer. 2022.
\newblock Selecting better samples from pre-trained llms: A case study on
  question generation.
\newblock \emph{arXiv preprint arXiv:2209.11000}.

\bibitem[{Zevenbergen and Whitehurst(2003)}]{zevenbergen2003dialogic}
Andrea~A Zevenbergen and Grover~J Whitehurst. 2003.
\newblock Dialogic reading: A shared picture book reading intervention for
  preschoolers.
\newblock \emph{On reading books to children: Parents and teachers}, 177:200.

\bibitem[{Zhang et~al.(2022)Zhang, Xu, Wang, Yao, Ritchie, Wu, Yu, Wang, and
  Li}]{zhang2022storybuddy}
Zheng Zhang, Ying Xu, Yanhao Wang, Bingsheng Yao, Daniel Ritchie, Tongshuang
  Wu, Mo~Yu, Dakuo Wang, and Toby Jia-Jun Li. 2022.
\newblock Storybuddy: A human-ai collaborative chatbot for parent-child
  interactive storytelling with flexible parental involvement.
\newblock In \emph{Proceedings of the 2022 CHI Conference on Human Factors in
  Computing Systems}, pages 1--21.

\bibitem[{Zhao et~al.(2022)Zhao, Hou, Wang, Yu, Liu, and
  Ma}]{zhao-etal-2022-educational}
Zhenjie Zhao, Yufang Hou, Dakuo Wang, Mo~Yu, Chengzhong Liu, and Xiaojuan Ma.
  2022.
\newblock \href {https://doi.org/10.18653/v1/2022.acl-long.348} {Educational
  question generation of children storybooks via question type distribution
  learning and event-centric summarization}.
\newblock In \emph{Proceedings of the 60th Annual Meeting of the Association
  for Computational Linguistics (Volume 1: Long Papers)}, pages 5073--5085,
  Dublin, Ireland. Association for Computational Linguistics.

\bibitem[{Ziegler et~al.(2019)Ziegler, Stiennon, Wu, Brown, Radford, Amodei,
  Christiano, and Irving}]{ziegler2019fine}
Daniel~M Ziegler, Nisan Stiennon, Jeffrey Wu, Tom~B Brown, Alec Radford, Dario
  Amodei, Paul Christiano, and Geoffrey Irving. 2019.
\newblock Fine-tuning language models from human preferences.
\newblock \emph{arXiv preprint arXiv:1909.08593}.

\bibitem[{Zou et~al.(2022)Zou, Li, Pan, and Aw}]{zou2022automatic}
Bowei Zou, Pengfei Li, Liangming Pan, and Aiti Aw. 2022.
\newblock Automatic true/false question generation for educational purpose.
\newblock In \emph{Proceedings of the 17th Workshop on Innovative Use of NLP
  for Building Educational Applications (BEA 2022)}, pages 61--70.

\bibitem[{Zylich et~al.(2020)Zylich, Viola, Toggerson, Al-Hariri, and
  Lan}]{zylich2020exploring}
Brian Zylich, Adam Viola, Brokk Toggerson, Lara Al-Hariri, and Andrew Lan.
  2020.
\newblock Exploring automated question answering methods for teaching
  assistance.
\newblock In \emph{Artificial Intelligence in Education: 21st International
  Conference, AIED 2020, Ifrane, Morocco, July 6--10, 2020, Proceedings, Part I
  21}, pages 610--622. Springer.

\end{thebibliography}



\clearpage

\appendix


\section*{Supplementary Material}
\label{sec:appendix}

\begin{algorithm}
 $\mathrm{T}\leftarrow \{(c_1, a_1, q_1), \ldots, (c_N, a_N, q_N)\}$\;
 \For{$i\leftarrow 1$ \KwTo $N$}{
  $\{\hat{q}_{i,1}, \ldots, \hat{q}_{i,M}\}\leftarrow$ GenQuesCodex(($c_i, a_i, q_i$))\;\label{line:gen_ques_codex}
  $\bar{a}_i\leftarrow$ GenAnsCodex($(c_i, q_i)$)\;\label{line:answer_codex_input_question}
  \For{$j\leftarrow 1$ \KwTo $M$}{
    $\hat{a}_{i,j}\leftarrow$ GenAnsCodex($(c_i, \hat{q}_{i,j})$)\;\label{line:answer_codex_generated_question}
    \If{$\text{ROUGE}(\hat{a}_{i,j}, a_i) > 0.5$ \textbf{or} $\text{ROUGE}(\hat{a}_{i,j}, \bar{a_i}) > 0.5$}{\label{line:consistency_check}
    $\mathrm{T}\leftarrow \mathrm{T} \cup \{(c_i, a_i, \hat{q}_{i,j})\}$\;
    }
  }
 }
\caption{Our automated data augmentation method which first generates question candidates and then filters them using consistency matching. 
}
\label{algo:data_augmentation}
\end{algorithm}

\begin{table}
\centering
\scalebox{0.95}{
\begin{tabular}{ll}
\toprule
Data Augmentation Method Variant &  ROUGE-L\\
\midrule

No Augmentation & 0.5639\\ 
All Questions & 0.5499\\
Minority Questions & 0.5607\\
Minority Questions + $\lambda$ Weighting & \textbf{0.5664}\\ 

\bottomrule
\end{tabular}}
\caption{Experimental results on the FairytaleQA dataset in ROUGE-L (higher is better) comparing different variants of our data augmentation method.}
\label{tab:aug_results}
\end{table}

\begin{table}
\centering
\small
\begin{tabular}{p{0.19\linewidth}p{0.19\linewidth}p{0.19\linewidth}p{0.19\linewidth}}

\toprule
Decoding Type & Perplexity-based Ranking & Distribution Matching-based Ranking with Nucleus Sampling (0.95, 1, 10) & Distribution Matching-based Ranking with Contrastive Search (4, 0.6, 10)\\ \midrule
Greedy (No ranking) & 0.5639 & 0.5639 & 0.5639 \\ 
\midrule
Nucleus Sampling (0.9, 1, 10) & 0.5664 & \textbf{0.5778} & 0.5657\\
\midrule
Nucleus Sampling (0.95, 1, 10) & 0.5618 & 0.5717 & \textbf{0.5678}\\
\midrule
Nucleus Sampling (0.95, 1, 75) & 0.5671 & 0.5766 & 0.5638\\
\midrule
Contrastive Search (4, 0.6, 10) & \textbf{0.5689} & 0.5719 & 0.5647\\ 
\bottomrule

\end{tabular}
\caption{Experimental results on the FairytaleQA dataset in ROUGE-L (higher is better) comparing different decoding strategies across our overgenerate-and-rank methods. We denote Nucleus Sampling N questions with a p value of P and temperature of T as Nucleus Sampling (P, T, N) and Contrastive Search of N questions with a top-k of K and alpha penalty of A as Contrastive Search (K, A, N).}
\label{tab:results_rank}
\end{table}

\begin{table*}
\centering
\small
\begin{tabular}{p{0.28\linewidth}p{0.12\linewidth}p{0.12\linewidth}p{0.1\linewidth}p{0.1\linewidth}p{0.1\linewidth}}
\toprule
Context & Ground Truth Answer & Ground Truth Question & Generated Question & Generated Answer of Generated Question & Generated Answer of Ground Truth Question\\ 
\midrule
\ldots and with that the rat laid a linen thread in the youth's hand. ``Heaven be praised!'', said the youth when he was up above once more. ``I'll not go down there again in a hurry.'' But he held the thread in his hand and danced and sang as usual \ldots 
& excited 
& How did the youth feel when the rat allowed him to go above? 
& How did the youth feel when he had the linen thread in his hand? & happy & happy\\ 
\bottomrule
\end{tabular}
\caption{Our data augmentation method on an example context-answer pair from FairytaleQA. We use two reference answers for consistency matching. In this example, although the generated answer of generated question (happy) does not match the reference ground truth answer (excited), the generated question is still consistent and included in the augmented training set since the generated answer matches the alternate reference of generated answer of the ground truth question (happy).}
\label{tab:gen_question}
\end{table*}

\begin{table*}
\small
\centering
\begin{tabular}{p{0.5\linewidth} p{0.09\linewidth}p{0.09\linewidth}p{0.09\linewidth}p{0.1\linewidth}}
\toprule
Context & Answer & Ground Truth Question & Generated Question & Error Type\\
\midrule
"What is your name?" asked the girl from underground. "Self is my name," said the woman. That seemed a curious name to the girl, and she once more began to pull
the fire apart. Then the woman grew angry and began to scold, and built
it all up again. Thus they went on for a good while; but at last, while
they were in the midst of their pulling apart and building up of the
fire, the woman upset the tar-barrel on the girl from underground. Then
the latter screamed and ran away, crying: "Father, father! Self burned me!" "Nonsense, if self did it, then self must suffer for it!" came the
answer from below the hill.
& The girl. & Who did the girl's father think burned the girl? & Who screamed and ran away? & Character coreference resolution\\

\midrule
So the gallows was built upon a high platform, and the fisher's son
mounted the steps up to it, and turned at the top to make the speech
that was expected from every doomed man, innocent or guilt. As he spoke
he happened to raise his arm, and the king's daughter, who was there at
her father's side, saw the name which she had written under it. With
a shriek she sprang from her seat, and the eyes of the spectators were
turned towards her. 'Stop! stop!' she cried, hardly knowing what she said. 'If that man
is hanged there is not a soul in the kingdom but shall die also.' And
running up to where the fisher's son was standing, she took him by the
hand, saying, 'Father, this is no robber or murderer, but the victor in the three
races, and he loosed the spells that were laid upon me.'
& The king's daughter saw the name which she had written under it. & How did the princess recognize Ian? & What happened when the fisher's son raised his arm? & Out-of-context ground-truth questions\\

\midrule
His vengeance was baulked, however, for in the panic and confusion that
followed Harold's death, the two Countesses slipped out of the Palace
and fled to the coast, and took boat in haste to Scotland, where they
had great possessions, and where they were much looked up to, and where
no one would believe a word against them. But retribution fell on them in the end, as it always does fall, sooner
or later, on everyone who is wicked, or selfish, or cruel; for the
Norsemen invaded the land, and their Castle was set on fire, and they
perished miserably in the flames. When Earl Paul found that they had escaped, he set out in hot haste for
the Island of Hoy, for he was determined that the Dwarf, at least,
should not escape. But when he came to the Dwarfie Stone he found it
silent and deserted, all trace of its uncanny occupants having
disappeared. 
& Norsemen invaded the land, and their Castle was set on fire, and they perished miserably in the flames. & What happened after the two Countesses fled to Scotland? & What happened because everyone who is wicked, or selfish, or cruel? & Multiple evidence angles in context\\
\bottomrule
\end{tabular}
\caption{Qualitative error analysis of our best method, distribution matching-based ranking, showing error examples from the FairytaleQA for each error type.}
\label{tab:qualitative_error_types}
\end{table*}

\end{document}